\documentclass[preprint,12pt]{elsarticle}

\usepackage{amssymb}
\usepackage{amsmath}

\usepackage{algorithmic}
\usepackage{algorithm}
\usepackage{hyperref}
\hypersetup{hidelinks=true}
\usepackage{textcomp}

\journal{Magnetic Resonance Imaging}

\begin{document}

\begin{frontmatter}



\title{Enhancing and Accelerating Brain MRI through Deep Learning Reconstruction Using Prior Subject-Specific Imaging}

\author[label1]{Amirmohammad Shamaei\corref{cor1}}
\ead{amirmohammad.shamaei@ucalgary.ca}
\cortext[cor1]{Corresponding author}

\author[label2,label4]{Alexander Stebner}
\author[label2,label5]{Salome (Lou) Bosshart}
\author[label2,label6,label6]{Johanna Ospel}
\author[label1]{Gouri Ginde}
\author[label1,label3,label6]{Mariana Bento}
\author[label1,label6]{Roberto Souza}

\affiliation[label1]{organization={Department of Electrical and Software Engineering, University of Calgary},
            city={Calgary},
            state={AB},
            country={Canada}}

\affiliation[label2]{organization={Departments of Radiology and Clinical Neuroscience, University of Calgary},
            city={Calgary},
            state={AB},
            country={Canada}}

\affiliation[label3]{organization={Department of Biomedical Engineering, University of Calgary},
            city={Calgary},
            state={AB},
            country={Canada}}

\affiliation[label4]{organization={Department of Radiology, University Hospital Basel},
            city={Basel},
            country={Switzerland}}

\affiliation[label5]{organization={Department of Neurology, University Hospital Basel},
            city={Basel},
            country={Switzerland}}

\affiliation[label6]{organization={Hotchkiss Brain Institute, University of Calgary},
            city={Calgary},
            state={AB},
            country={Canada}}

\begin{abstract}
Magnetic resonance imaging (MRI) is a crucial medical imaging modality. However, long acquisition times remain a significant challenge, leading to increased costs, and reduced patient comfort. Recent studies have shown the potential of using deep learning models that incorporate information from prior subject-specific MRI scans to improve reconstruction quality of present scans. Integrating this prior information requires registration of the previous scan to the current image reconstruction, which can be time-consuming. We propose a novel deep-learning-based MRI reconstruction framework which consists of an initial reconstruction network, a deep registration model, and a transformer-based enhancement network.  We validated our method on a longitudinal dataset of T1-weighted MRI scans with 2,808 images from 18 subjects at four acceleration factors (R5, R10, R15, R20). Quantitative metrics confirmed our approach's superiority over existing methods (p $<$ 0.05, Wilcoxon signed-rank test). Furthermore, we analyzed the impact of our MRI reconstruction method on the downstream task of brain segmentation and observed improved accuracy and volumetric agreement with reference segmentations. Our approach also achieved a substantial reduction in total reconstruction time compared to methods that use traditional registration algorithms, making it more suitable for real-time clinical applications. The code associated with this work is publicly available at \url{https://github.com/amirshamaei/longitudinal-mri-deep-recon}.
\end{abstract}


\end{frontmatter}

\section{Introduction}
\label{sec:introduction}

Brain magnetic resonance (MR) imaging is an essential radiation-free medical imaging modality that provides unparalleled soft tissue contrast \cite{Tourais2022}. Its non-invasive nature and versatility have made it a crucial tool for patient diagnostics, monitoring and neurosciences research. However, the prolonged acquisition times associated with MR imaging pose significant challenges, including increased operational costs, reduced patient throughput, and lengthy wait times \cite{bhandari2017radiology}. These issues not only strain healthcare resources but also contribute to patient discomfort and potential motion artifacts, which can degrade image quality.

In recent years, various strategies have been proposed to accelerate MR acquisitions while maintaining diagnostic image quality. Parallel imaging (PI) techniques, such as SENSE \cite{sense} and GRAPPA \cite{garppa}, exploit the spatial information from multiple receiver coils to reduce the number of required phase-encoding steps \cite{Deshmane2012, Cummings2022}. Compressed sensing (CS) approaches leverage the inherent sparsity of MR images to undersample $k$-space data, resulting in faster acquisitions \cite{Lustig2007}. While these methods have shown promise, they often require complex optimization algorithms and may introduce artifacts in the reconstructed images \cite{Liu2022}.

The advent of deep learning has opened new avenues for accelerating MR image reconstruction \cite{Hammernik2018,Hammernik2022}. Deep neural networks have demonstrated remarkable ability in learning the complex mapping between undersampled $k$-space data and fully-sampled reference images \cite{Heckel2024}. By training on large datasets, these models can effectively capture the underlying structure and redundancies in MR images, enabling high-quality reconstruction from heavily undersampled data. This has the potential to significantly reduce acquisition times without compromising diagnostic value.

A recent study by Souza et al. \cite{souza2020enhanced} introduced a novel framework that incorporates prior subject-specific MR images to enhance the reconstruction quality of current scans. This approach leverages the inherent similarity between longitudinal MR images of the same subject, which are often readily available through the picture archiving and communication systems (PACS). By registering the previous scan to the initial reconstruction of the current scan and using an enhancement network, they obtained image reconstruction metrics for $R=15$ superior to $R=5$ compared to non-enhanced methods. Leveraging prior subject-specific MR imaging sessions can also enhance motion correction \cite{Chatterjee2020}.

A key limitation of Souza et al.'s approach \cite{souza2020enhanced} is the reliance on linear registration techniques to align the prior and current scans. While effective, linear registration can be computationally intensive and time-consuming, hindering the practical implementation of this method in clinical settings where rapid reconstruction is crucial. Moreover, linear registration may not adequately capture the complex, non-linear deformations that can occur between scans due to factors such as patient positioning, anatomical changes, and disease progression. Also,  while the enhancement network proposed by Souza et al. demonstrated promising results, there is potential for further improvement in capturing complex spatial relationships and long-range dependencies within MR images.

To address these issues, we propose an extension of Souza et al.'s work \cite{souza2020enhanced} that incorporates a state-of-the-art deep learning technique for image registration and a transformer-based enhancement network \cite{dosovitskiy2020image}.   Our method leverages a deep-learning-based registration tool to efficiently align a prior scan with the initial reconstruction, improving the accuracy and speed of the alignment process.  This accurately registered prior information is then used to enhance the reconstruction through a transformer-based network. Transformers' self-attention mechanisms allow for efficient modelling of global context and relationships between distant image regions, which is particularly beneficial for capturing subtle anatomical details and preserving structural integrity. By leveraging the strengths of transformer architectures, we aim to enhance the ability of the reconstruction framework to integrate prior subject-specific information and produce higher-quality reconstructions, especially in cases where traditional convolutional approaches may struggle to capture intricate spatial relationships. Our proposed framework aims to generate high-quality reconstructions while significantly reducing the total reconstruction time compared to traditional methods that rely on computationally expensive linear registration techniques.

To validate our proposed method, we conducted extensive experiments using a dataset of longitudinally acquired, three-dimensional, T1-weighted brain images with varying acquisition parameters. We assessed the reconstruction quality using quantitative metrics such as structural similarity (SSIM), peak signal-to-noise ratio (PSNR), and normalized root mean squared error (NRMSE). Additionally, we analyzed the impact of our fast MRI reconstruction method on the downstream task of brain segmentation and observed improved accuracy and volumetric agreement with reference segmentations.

The main contributions of this work are summarized as follows:
\begin{itemize}

\item[1.] We proposed to use a fast and accurate deep-learning-based registration method to efficiently align prior subject-specific brain imaging data with the initial reconstruction, overcoming the limitations of linear registration techniques and providing computational speed-ups.

\item[2.] We introduced a transformer-based architecture \cite{Vaswani2017} in the enhancement network to capture long-range dependencies and improve reconstruction accuracy, leveraging the power of self-attention mechanisms.

\item[3.] We conducted a comprehensive evaluation of our method using a longitudinal dataset and rigorous analysis pipeline, demonstrating significant improvements in reconstruction quality compared with previous approaches and non-enhanced reconstruction models.

\item[4.] We analyzed the effect of fast MRI reconstruction on the downstream task of brain segmentation.
\end{itemize}

The remainder of this paper is organized as follows: Section 2 provides a detailed description of the materials and methods used in this study, including the dataset, proposed processing model, and experimental setup. Section 3 presents the results of our experiments and discusses the implications of our findings, comparing the performance of our proposed method with non-enhanced and linearly registered enhanced reconstructions. Section 4 discusses the impact and limitations of our method. Finally, Section 5 concludes the paper, highlighting the significance of our contributions and outlining future research directions.

\begin{figure*}[!t]
\centerline{\includegraphics[width=0.85\textwidth]{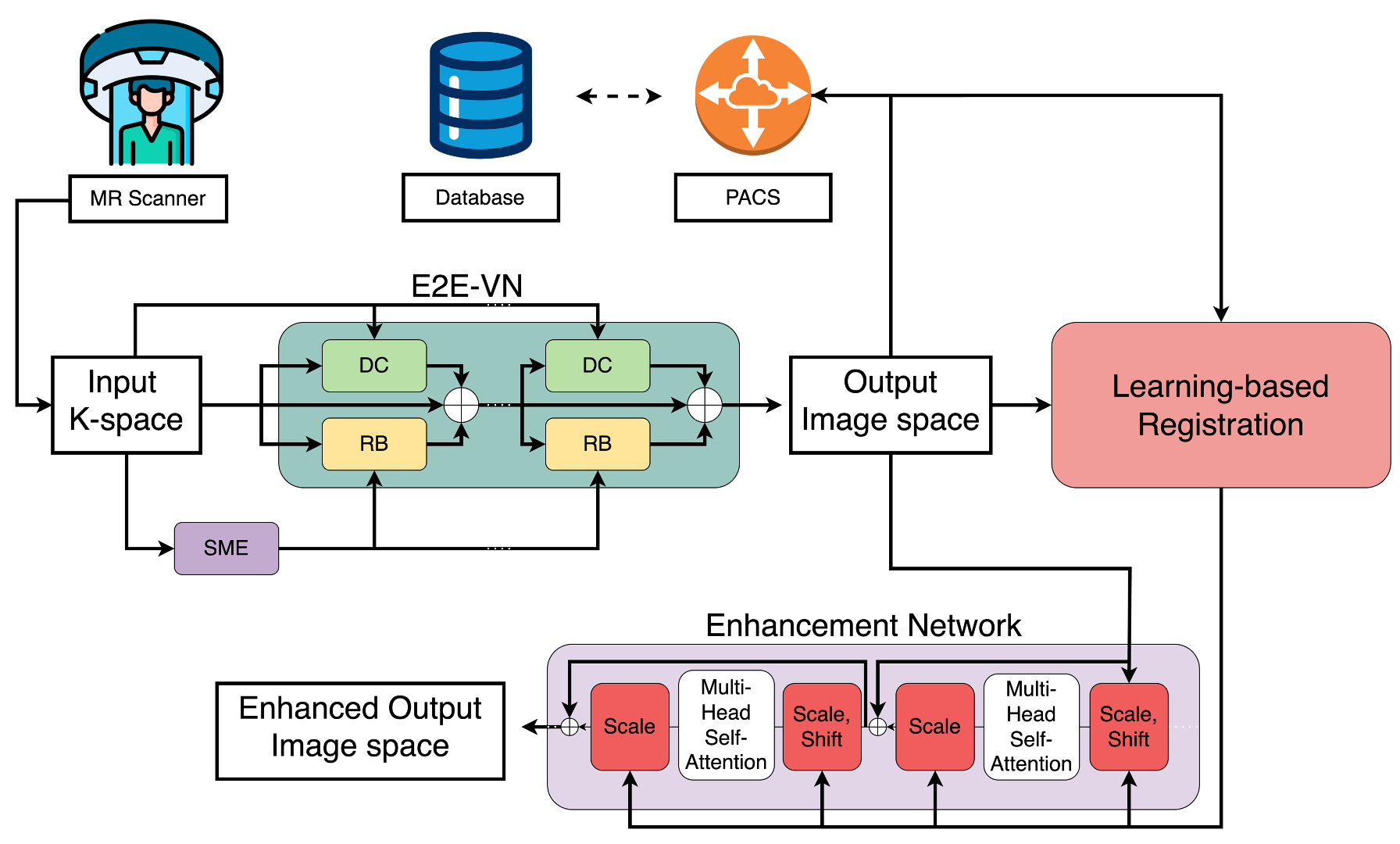}}
\caption{Overview of the proposed MR image reconstruction pipeline. The undersampled $k$-space data is initially reconstructed using the E2E-VarNet network. The reconstructed image is then aligned with prior subject-specific data using a deep-learning-based registration method (EasyReg). The registered prior data is fed into a transformer-based enhancement network, which utilizes multi-head self-attention and scaling/shifting operations to refine the reconstructed image. The enhanced output image space serves as the final high-quality reconstruction result.}
\label{fig:figure1}
\end{figure*}

\section{Materials and Methods}
\subsection{Dataset}
In this study, we utilized the same dataset as Souza et al. \cite{souza2020enhanced}, which consists of three-dimensional (3D), T1-weighted, gradient-recalled echo, sagittal acquisitions obtained from a 3T MR scanner (Discovery MR750, General Electric (GE) Healthcare, Waukesha, WI) from 79 subjects. Among these images, 39 have a corresponding fully sampled reconstructed volume from a previous time point available in DICOM format \cite{Mildenberger2002}, resulting in 39 pairs of longitudinal exams. The imaging data were acquired using a 12-channel imaging coil from presumably healthy subjects with a mean age of 45 years ± 16 years (range: 20 years to 80 years). The acquisition parameters were TR/TE/TI = 6.3 ms/2.6 ms/650 ms or 7.4 ms/3.1 ms/400 ms, leading to slight variations in image contrast and increased dataset heterogeneity. The mean time interval between scans was 4 years ± 1.17 years (range: 2.95 years to 6.14 years), and given the older subject population, changes due to normal aging are anticipated.

The acquisitions employed a field of view of 256 mm $\times$ 218 mm, collecting 170 to 180 contiguous 1.0-mm thick images. The acquisition matrix for each channel was $N_x \times N_y \times N_z = 256 \times 218 \times [170, 180]$. To reduce scan time to approximately 6 minutes, partial-encoding (85\%) along $k_z$ was used, which is a common practice in both clinical and research MR imaging. The scanner automatically applied the inverse Fourier Transform (FT) to the $k$-space data in the frequency-encoded direction ($k_x$), resulting in a hybrid $x - k_y - k_z$ data set. This initial transform simplified the reconstruction task from a 3D to a 2D problem, allowing undersampling to occur in the phase encoding ($k_y$) and slice encoding ($k_z$) directions. Lastly, the fully sampled reconstructed images (reference images) were obtained by applying the channel-wise inverse 2D FT to the hybrid $x - k_y - k_z$ data and combining the channels using standard root sum-of-squares processing.

\subsection{Proposed Processing Model}
Our proposed processing model extends the work of Souza et al. \cite{souza2020enhanced} by incorporating a deep-learning-based registration method and a transformer-based enhancement network. The pipeline consists of the following key components:
\begin{itemize}

\item[1.] Initial Reconstruction: We employed the end-to-end variational network (E2E-VarNet) \cite{Sriram2020} for the initial reconstruction of the undersampled $k$-space data. E2E-VarNet is a state-of-the-art deep learning model that learns to estimate sensitivity maps and performs reconstruction in a unified framework.

\item[2.] Deep Registration: To align the prior subject-specific brain imaging data with the initial reconstruction, we utilized the EasyReg model\cite{Iglesias2023}, a fast and accurate deep-learning-based registration tool. EasyReg employs a convolutional neural network to learn the optimal transformation parameters, enabling non-linear registration in a computationally efficient manner.

\item[3.] Enhancement Network: We introduced a transformer-based architecture in the enhancement network to capture long-range dependencies and improve reconstruction accuracy \cite{Peebles2023,Vaswani2017}. The transformer model, inspired by the success of self-attention mechanisms in natural language processing and computer vision tasks, allows for efficient modeling of global context and relationships between image regions.

\end{itemize}

Figure \ref{fig:figure1} presents an overview of the proposed deep-learning-based MRI reconstruction framework. The process begins with the acquisition of undersampled $k$-space data using an MR scanner. The undersampled $k$-space data is then fed into the E2E-VarNet \cite{Sriram2020}, which consists of three main components: a data consistency (DC) block, a refinement block (RB), and a sensitivity map estimation (SME) block. The DC block ensures that the reconstructed image remains consistent with the acquired $k$-space data, while the R block maps multi-coil $k$-space data into one image, applies a U-Net, and then transforms the data back to multi-coil $k$-space data. The SME block estimates the sensitivity maps of the receiver coils, which then are used in the R block.
The output image space from the E2E-VarNet network, referred to as the non-enhanced reconstruction, is then divided into patches, similar to the Vision Transformer (ViT) approach \cite{dosovitskiy2020image}. These patches are fed into a transformer-based enhancement network that utilizes multi-head self-attention mechanisms \cite{Vaswani2017, Peebles2023}. The self-attention allows the network to capture long-range dependencies and improve the reconstruction quality.
In parallel, a previous subject-specific brain image is retrieved from a picture archiving and communication system (PACS) database. This previous image is registered to the non-enhanced reconstruction using a learning-based registration method, such as EasyReg \cite{Iglesias2023}. The registration process aligns the previous image with the current reconstruction, enabling the enhancement network to leverage the prior anatomical information.
The registered previous image is then patched, converted into embedding vectors, and positionally encoded. These embedding vectors are used to shift and scale the outputs of the multi-head self-attention in the enhancement network. By incorporating the prior subject-specific information through the shifting and scaling operations, the enhancement network can produce a more accurate and detailed reconstruction of the undersampled MRI data.

\subsection{Mathematical Representation}
Let $\mathbf{X} \in \mathbb{C}^{N_y \times N_z \times N_c}$ represent the fully-sampled $k$-space data, where $N_y$, $N_z$, and $N_c$ denote the number of phase-encoding steps, slice-encoding steps, and coils, respectively. The fully-sampled reconstructed image $\mathbf{Y} \in \mathbb{R}^{N_y \times N_z}$ is obtained by applying the inverse Fourier transform $\mathcal{F}^{-1}$ to the $k$-space in each coil in  $\mathbf{X}$ and combining the coil images using the root-sum-of-squares (RSS) method:
\begin{equation}
\mathbf{Y} = \text{RSS}(\mathcal{F}^{-1}(\mathbf{X}))
\end{equation}
The undersampled $k$-space data $\mathbf{X}_u \in \mathbb{C}^{N_y \times N_z \times N_c}$ is obtained by element-wise multiplication of the fully-sampled $k$-space data $\mathbf{X}$ with the binary undersampling mask $\mathbf{M} \in \{0, 1\}^{N_y \times N_z \times N_c}$:
\begin{equation}
\mathbf{X}_u = \mathbf{M} \odot \mathbf{X}
\end{equation}
where $\odot$ represents the element-wise multiplication operator.
The initial reconstruction $\hat{\mathbf{Y}}$ is obtained by applying the E2E-VarNet model, $f_\theta$, to the undersampled $k$-space data $\mathbf{X}_u$:
\begin{equation}
\hat{\mathbf{Y}} = f_\theta(\mathbf{X}_u)
\end{equation}
where $\theta$ denotes the learnable parameters of the E2E-VarNet model.

The previous scan $\mathbf{PS} \in \mathbb{R}^{N_y \times N_z}$ is registered to the initial reconstruction $\hat{\mathbf{Y}} \in \mathbb{R}^{N_y \times N_z}$ using the deep registration model $g_\phi$:

\begin{equation}
\mathbf{PS}_\text{reg} = g_\phi(\mathbf{PS}, \hat{\mathbf{Y}})
\end{equation}
where $\phi$ represents the learnable parameters of the deep registration model.
The enhanced reconstruction $\hat{\mathbf{Y}}_\text{enh}$ is obtained by applying the transformer-based enhancement network $h_\psi$ to the concatenation of the initial reconstruction $\hat{\mathbf{Y}}$ and the registered previous scan $\mathbf{P}_\text{reg}$:
\begin{equation}
\hat{\mathbf{Y}}_\text{enh} = h_\psi(\hat{\mathbf{Y}}, \mathbf{PS}_\text{reg})
\end{equation}
where $\psi$ denotes the learnable parameters of the transformer-based enhancement network.
The loss function $\mathcal{L}$ used for training the proposed deep learning framework is defined as  the structural similarity index (SSIM) loss between the enhanced reconstruction $\hat{\mathbf{Y}}_\text{enh}$ and the fully-sampled reference image $\mathbf{Y}$, both of which are first normalized by their absolute maximum values.
\begin{equation}
\mathcal{L}(\hat{\mathbf{Y}}_\text{enh}, \mathbf{Y}) =  (1 - \text{SSIM}(\hat{\mathbf{Y}}_\text{enh}, \mathbf{Y}))
\end{equation}
The learnable parameters ${\theta, \phi, \psi}$ of the proposed deep learning framework are optimized using the Adam optimizer to minimize the loss function $\mathcal{L}$ over the training dataset.

\subsection{Experimental setup}
The experiments were designed to compare the performance of our approach with non-enhanced reconstruction and linearly registered enhanced reconstruction using FSL FLIRT \cite{jenkinson2001global}, as used in Souza et al.'s work \cite{souza2020enhanced}. We also assessed the impact of using a transformer versus using a UNet architecture in the image enhancement block. Table \ref{tab:image_enhancement_comparison} summarizes the different model configurations compared.

\begin{table}[!t]
\caption{Summary of model components and processing times.}
\centering
\label{tab:image_enhancement_comparison}
\begin{tabular}{p{2.5cm}|p{2.2cm}|p{2.2cm}|p{2.2cm}|p{0.8cm}|p{0.8cm}}
\hline
Method & Initial Rec. & Reg. & Enhanced Rec. & Reg. (s) & Rec. (s)\\
\hline
Enhanced [Ours, Transformers] & E2EVarNet & EasyReg & Transformer & 4.1 & 3.42\\
\hline
Enhanced [Ours, UNet] & E2EVarNet & EasyReg & UNet & 4.1 & 1.21\\
\hline
Enhanced [Souza et al., UNet] & WW-net IKIK & FSL FLIRT & UNet & 90 & 3.10\\
\hline
Non-enhanced [Ours, E2EVarNet] & E2EVarNet & - & - & - & 1.10\\
\hline
Non-enhanced [Souza et al., WW-net IKIK] & WW-net IKIK & - & - & - & 2.78 \\
\hline
\end{tabular}
\end{table}

In our study, we trained reconstruction and enhancement models for four different acceleration factors (R5, R10, R15, R20) using retrospective undersampling. One E2E-VarNet \cite{Sriram2020} was trained for each acceleration factor for the non-enhanced reconstruction. We employed the same Poisson disc distribution sampling scheme as Souza et al. \cite{souza2020enhanced}, with the center of $k$-space fully sampled within a circle of radius 16. To ensure a fair comparison between the non-enhanced and enhanced reconstruction models, we utilized the same network capacity for both models.

The non-enhanced and enhanced reconstruction networks were trained for 200 and 100 epochs using the Adam optimizer with a initial learning rate of $1\times10^{-3}$, and a batch size of 64 and 32, respectively. For the enhancement network, we employed data augmentation techniques such as rotation up to 15°, vertical and horizontal translation up to 10\% of the image dimensions, and scaling up to 10\%. To further improve the training process, we implemented a learning rate adjustment strategy that reduces the learning rate when the model's performance on the validation set stops improving. To prevent the models from overfitting to the training data and to ensure optimal performance, we employed an early stopping technique. If the model's performance on the validation accuracy metric (SSIM) did not improve for 10 consecutive epochs, the training process was stopped.

For training and validating the E2E-VarNet model, we used 43 and 18 subjects, respectively. Each subject had a volumetric acquisition with Nx = 256. The peripheral 50 slices from each end of the volumetric data were excluded, resulting in 6,708 images (slices) available for training and 2,808 images for validation. For the enhancement network, we used 15 subjects (2,340 images) for training and 6 subjects (936 images) for validation. All subjects used for training and validation of the enhancement network had previous scans available. The performance metrics for the non-enhanced and enhanced reconstructions were reported on an independent test set of 18 subjects (2,808 images), all of whom had previous scans. We split the dataset into training, validation, and testing subsets, ensuring that scans from the same subject were not present in different subsets to avoid data leakage. The E2E-VarNet model and the transformer-based enhancement network were trained using the SSIM loss function, optimizing for perceptual quality. The models were evaluated on the held-out testing set using quantitative metrics such as SSIM, PSNR and NRMSE. These metrics were computed against the fully sampled current scan reference image using the scikit-image implementation. Higher SSIM and pSNR values and lower NRMSE values indicate better reconstruction performance

To assess the impact of our fast MRI reconstruction approach on downstream tasks, we performed segmentation analysis using the SynthSeg \cite{Billot2023} tool from FreeSurfer \cite{Fischl2012}. SynthSeg is a deep learning-based segmentation algorithm that automatically labels brain structures in MR images. We also visually inspected the FreeSurfer segmentation masks for quality control. The segmentation results were compared with those obtained from the fully-sampled reference images using the  Dice similarity coefficient (DSC) and volumetric agreement. Statistical significance was assessed using the Wilcoxon signed-rank test with an $\alpha$ of 0.05.

To further test the robustness and generalizability of our proposed method, we conducted two ablation studies using the ICBM 152 extended nonlinear atlas \cite{Fonov2009,Fonov2011} as a substitute for subject-specific previous scans. The ICBM 152 atlas\cite{Fonov2009,Fonov2011} is widely used in neuroimaging studies and is representative of a diverse population, making it a suitable choice for evaluating the generalizability of your method. In the first ablation study, we evaluated the performance of the transformer-based model from our primary experiments, which had been trained to leverage subject-specific prior scans. For this test, however, we replaced the subject-specific previous scan with the ICBM 152 atlas \cite{Fonov2009,Fonov2011}, which was registered to the initial reconstruction. This experiment aimed to assess the model's ability to leverage prior anatomical information from a standardized atlas instead of relying on subject-specific data. In the second ablation study, we trained our model directly using the registered atlas images and tested it using the same atlas in place of the previous scan. This approach allowed us to investigate the potential of using a standardized atlas as a consistent prior for enhancing MRI reconstruction quality, eliminating the need for subject-specific previous scans.

\subsection{Expert Readers Image Quality Assessment}

A board-certified radiologist and a neurology resident independently evaluated the diagnostic quality of reconstructed MRI images in this study. The assessment included images from six subjects, comparing non-accelerated scans with those obtained at four acceleration factors (R5, R10, R15, R20). The raters were blinded to the acceleration status of the images. The evaluation criteria, rated on a 4-point Likert scale (1-4), encompassed overall image quality, contrast and tissue differentiation, and preservation of fine anatomical details.

\section{Results}
\begin{figure*}[!t]
\centering{\includegraphics[width=0.95\textwidth]{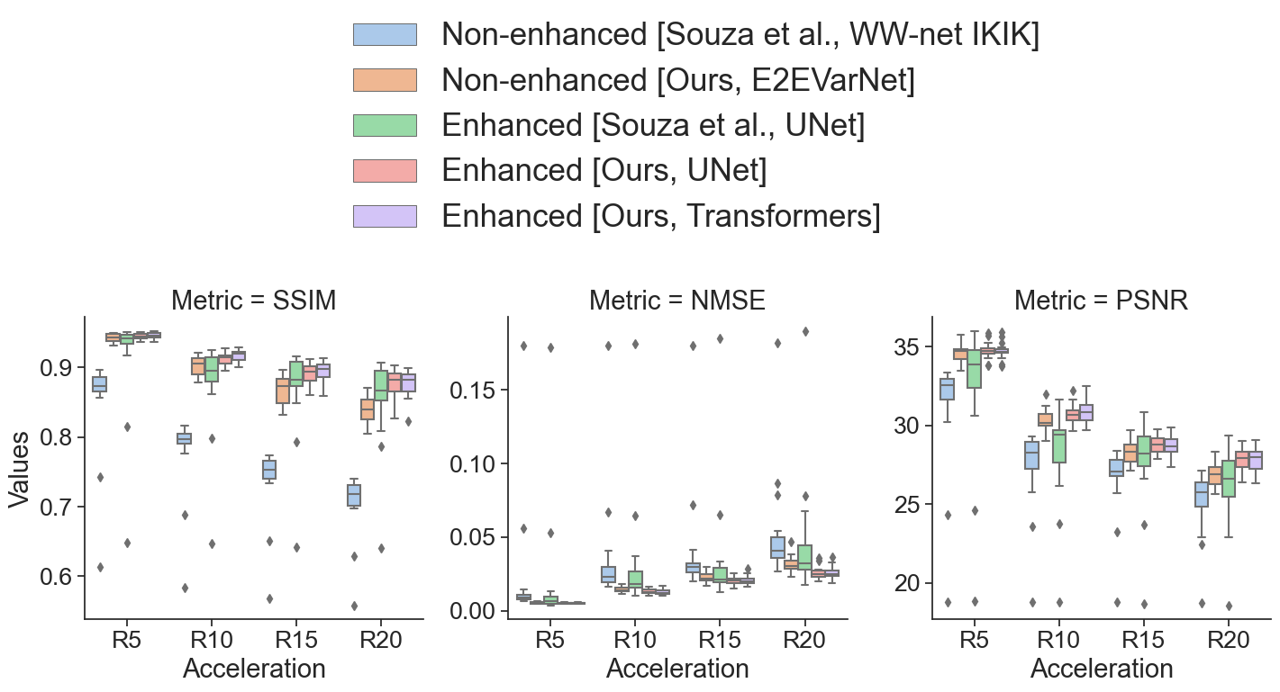}}
\caption{Comparison of the reconstruction quality of different methods across various acceleration factors (R5, R10, R15, R20) using three quantitative metrics: structural similarity (SSIM), normalized mean squared error (NMSE), and peak signal-to-noise ratio (PSNR). The enhanced reconstructions using the proposed transformer-based approach (Enhanced [Ours, Transformers]) consistently outperform the non-enhanced reconstructions (Non-enhanced [Ours, E2EVarNet] and Non-enhanced [Souza et al., WW-net IKIK]) and the previously proposed UNet-based enhanced reconstruction (Enhanced [Souza et al., UNet]) across all acceleration factors and evaluation metrics, demonstrating the superiority of the proposed method in preserving image quality while enabling faster MRI acquisitions.}
\label{fig:figure2}
\end{figure*}

\begin{figure*}[ht]
\centering{\includegraphics[width=0.7\textwidth]{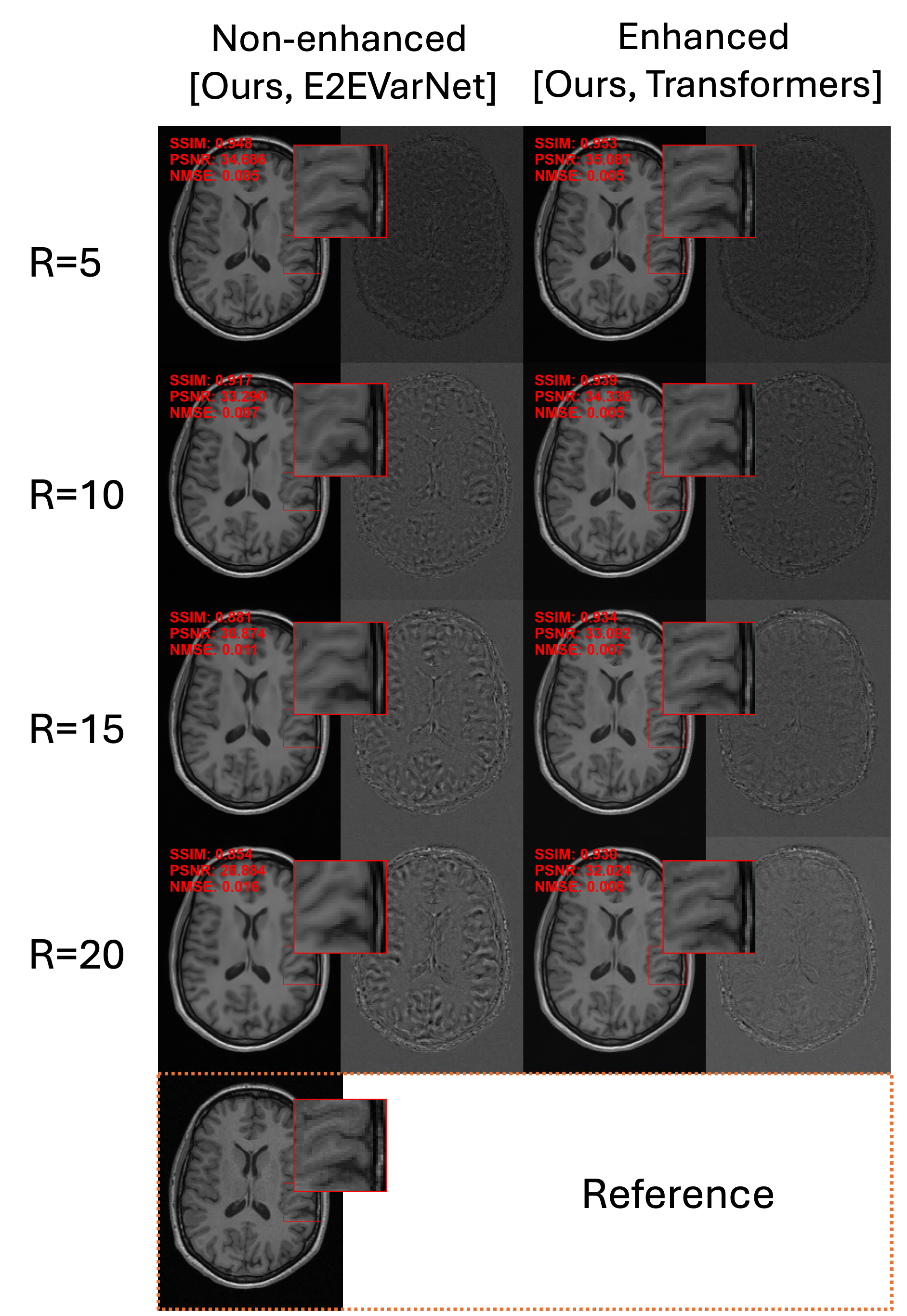}}
\caption{Visual comparison of the reconstructed brain MR images using the proposed transformer-based enhanced reconstruction approach (Enhanced [Ours, Transformers]) and the non-enhanced reconstruction (Non-enhanced [Ours, E2EVarNet]) against the fully-sampled reference image across different acceleration factors (R5, R10, R15, R20). The enhanced reconstructions exhibit superior image quality, preserving fine anatomical details and reducing artifacts, particularly at higher acceleration factors. The proposed method demonstrates its ability to maintain high perceptual similarity to the reference image while enabling faster MRI acquisitions, highlighting its potential for clinical adoption in accelerating MR examinations without compromising diagnostic quality.}
\label{fig:figure3}
\end{figure*}

The performance of our proposed methods, {Enhanced [Ours, UNet]} and {Enhanced [Ours, Transformers]}, were evaluated against the non-enhanced and linearly registered enhanced reconstructions from Souza et al.'s work \cite{souza2020enhanced}. Table \ref{tab:image_enhancement_comparison} summarizes the processing times for each of the compared methods, and Figure \ref{fig:figure2} shows the quantitative results for different acceleration factors in terms of SSIM, PSNR, and NMSE metrics. For the lowest acceleration factor (R=5), our transformer-based enhanced reconstruction achieved the highest SSIM of 0.9457 and PSNR of 34.73 dB, while maintaining a low NMSE of 0.0051. These values outperformed both the E2EVarNet non-enhanced reconstruction and the linearly registered enhanced reconstruction from Souza et al.'s \cite{souza2020enhanced}, demonstrating the efficacy of our deep registration and transformer-based enhancement network.
As the acceleration factor increased, the performance gap between our transformer-based method and the previous approaches became more pronounced. At an acceleration factors of 10, our enhanced reconstruction achieved an SSIM of 0.9171, PSNR of 30.82 dB, and NMSE of 0.0127, significantly better than the non-enhanced and linearly registered enhanced reconstructions. Similar trends were observed for higher acceleration factors of 15 and 20, with our methods consistently outperforming the others. Figure \ref{fig:figure3} illustrates the visual quality of the reconstructed images for different acceleration factors and methods against the fully-sampled reference image. Our transformer-based enhanced reconstruction (Enhanced [Ours, Transformers]) exhibits superior image quality, preserving fine details and structural information compared to the non-enhanced reconstructions (Non-enhanced [Ours, E2EVarNet]).

\begin{figure*}[ht]
\centering
\includegraphics[width=0.6\textwidth]{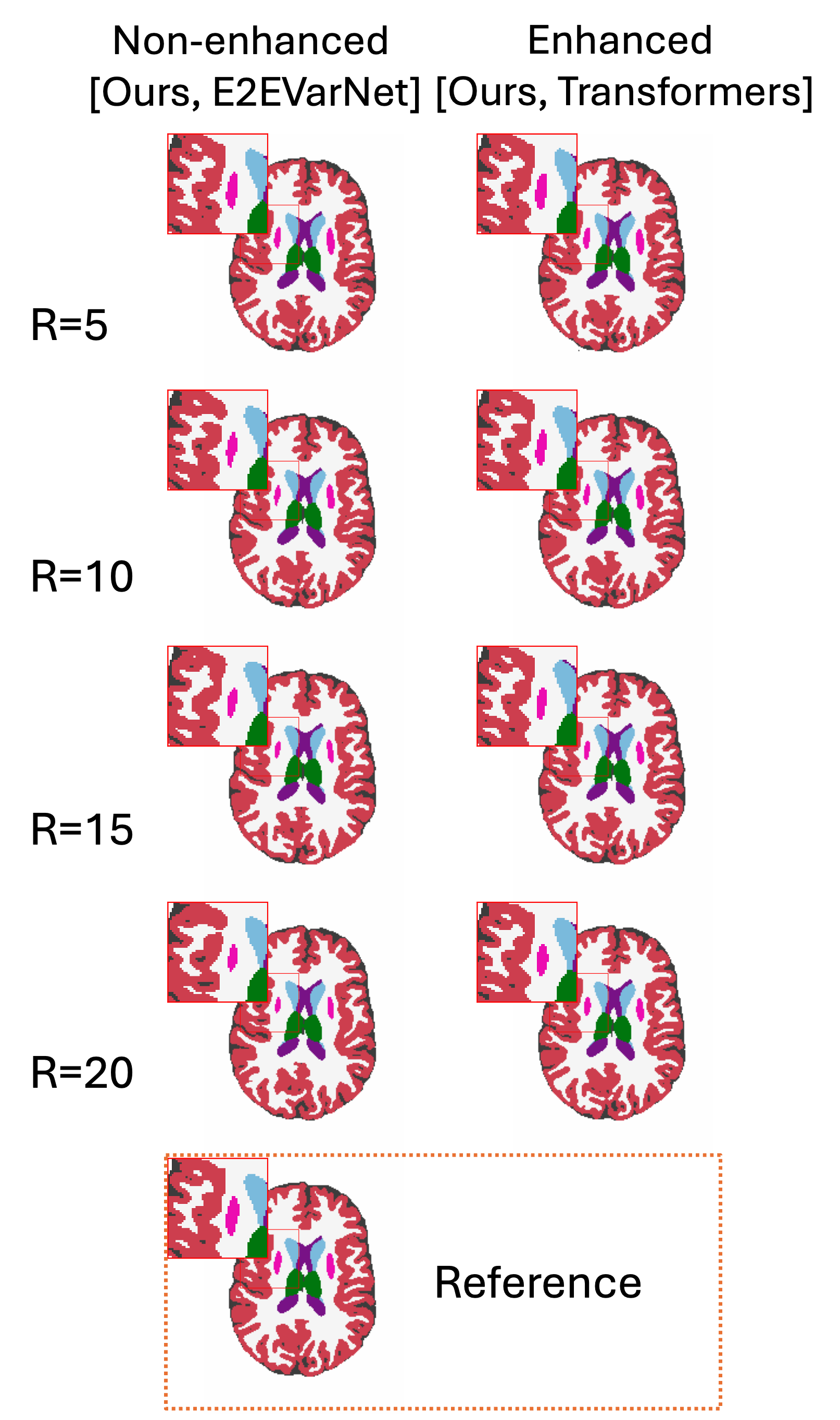}
\caption{Visual comparison of the reconstructed images using the proposed transformer-based enhanced reconstruction (Enhanced [Ours, Transformers]), non-enhanced reconstruction (Non-enhanced [Ours, E2EVarNet]), and the fully-sampled reference image for an example subject. The segmentation masks derived from the enhanced reconstruction exhibit higher agreement with the reference, particularly in the cortical regions, compared to the non-enhanced reconstruction, demonstrating the superior quality of the proposed method in preserving fine structural details crucial for accurate segmentation.}
\label{fig:figure4}
\end{figure*}

Figure \ref{fig:figure4} shows the segmentation masks obtained from our enhanced reconstruction (Enhanced [Ours, Transformers]), non-enhanced reconstruction [Ours, E2EVarNet], and the fully-sampled reference image for an example subject. The segmentation masks derived from our enhanced reconstruction exhibit higher agreement with the reference, particularly in the cortical regions, compared to the non-enhanced reconstruction.

\begin{figure}[t]
\centering
\includegraphics[width=\columnwidth]{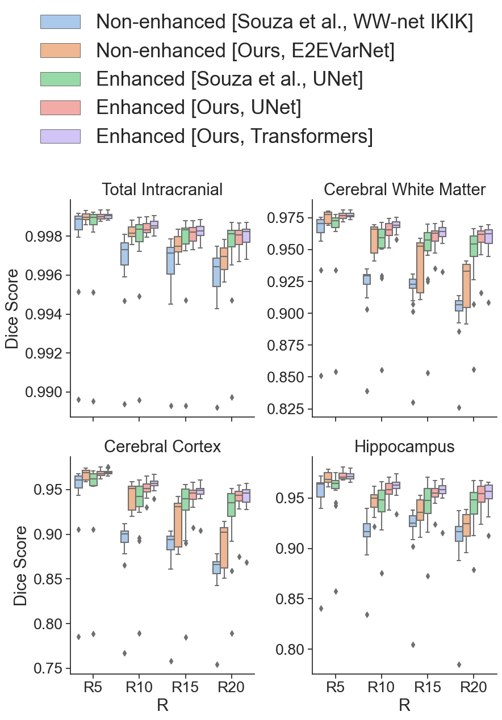}
\caption{Dice score coefficients for various brain regions (total intracranial, cerebral white matter, cerebral cortex, and hippocampus) across different acceleration factors (R5, R10, R15, R20). The proposed transformer-based enhanced reconstruction (Enhanced [Ours, Transformers]) consistently achieves higher Dice scores compared to the non-enhanced and previously proposed enhanced reconstructions, demonstrating its superiority in preserving the structural integrity of brain regions.}
\label{fig:figure5}
\end{figure}

Quantitative analysis of the segmentation results further confirmed the benefits of our approach. Figure \ref{fig:figure5} depicts the Dice similarity coefficients (DSC) for various brain regions across different acceleration factors. Our enhanced reconstruction consistently achieved higher DSC values, indicating better agreement with the reference segmentation.

\begin{figure}[t]
\centering
\includegraphics[width=\columnwidth]{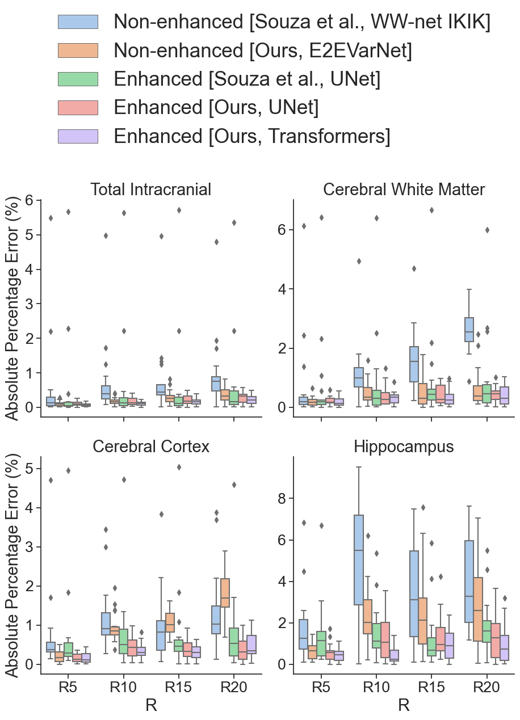}
\caption{Absolute percentage error in volume estimates for different brain regions (total intracranial, cerebral white matter, cerebral cortex, and hippocampus) across various acceleration factors (R5, R10, R15, R20). The proposed transformer-based enhanced reconstruction (Enhanced [Ours, Transformers]) exhibits lower percentage errors compared to the non-enhanced and previously proposed enhanced reconstructions, highlighting its ability to maintain accurate volumetric measurements despite faster MRI acquisitions.}
\label{fig:figure6}
\end{figure}

Furthermore, we evaluated the volumetric agreement between the segmented brain regions from the reconstructed images and the fully-sampled reference. Figure \ref{fig:figure6} shows the percent difference in estimated volumes for the brain, white matter, gray matter, and cortex. Our enhanced reconstruction exhibited the smallest deviations from the reference volumes across all acceleration factors, highlighting the potential impact of our method on clinical and research applications that rely on accurate volumetric measurements.

In terms of computational efficiency, our deep registration approach using EasyReg significantly reduced the registration time compared to the linear registration method employed by Souza et al \cite{souza2020enhanced}. The average registration time for our method was 4.1 seconds per subject, while the linear registration implemented in FSL FLIRT took approximately 90 seconds. This substantial reduction in registration time, combined with the improved reconstruction quality, makes our approach more feasible for real-time clinical applications.

\begin{figure*}[!t]
\centering{\includegraphics[width=0.8\textwidth]{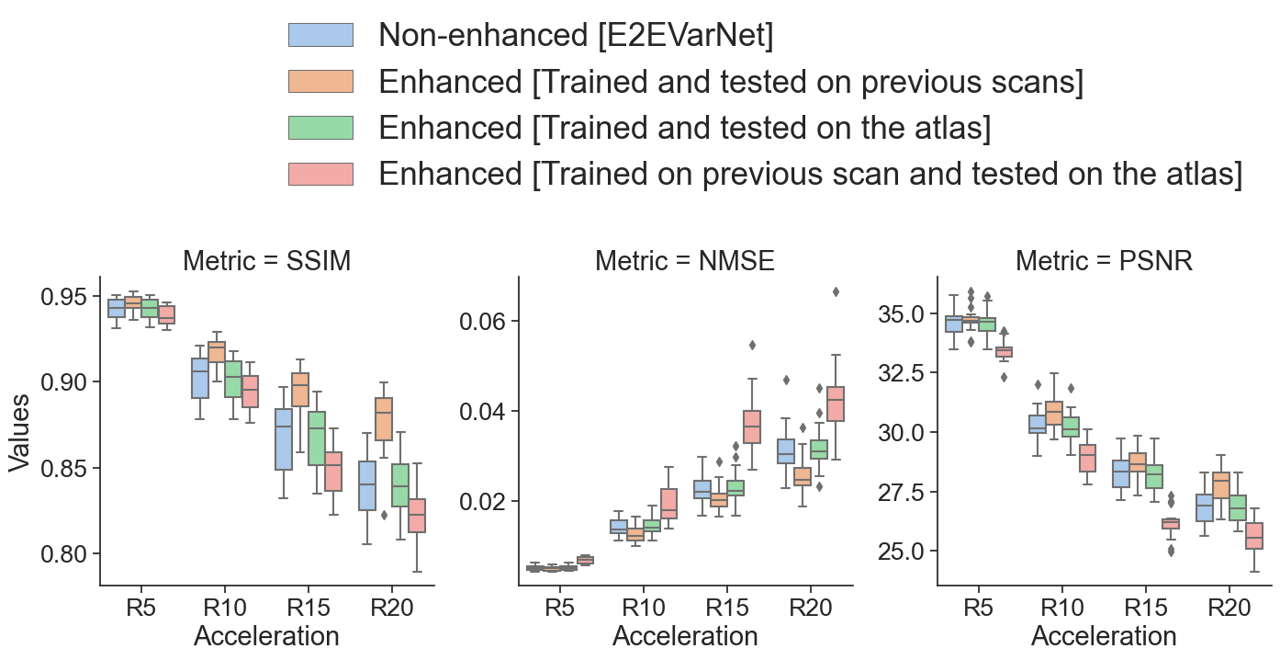}}
\caption{Comparison of the reconstruction quality of different methods across various acceleration factors (R5, R10, R15, R20) using three quantitative metrics: structural similarity (SSIM), normalized mean squared error (NMSE), and peak signal-to-noise ratio (PSNR). The enhanced reconstructions using the proposed transformer-based approach (Enhanced [Ours, Transformers]) consistently outperform the non-enhanced reconstructions (Non-enhanced [Ours, E2EVarNet] and enhanced reconstruction using an atlas across all acceleration factors and evaluation metrics.}
\label{fig:figure_}
\end{figure*}

Figure \ref{fig:figure_} presents the results of these ablation studies, comparing the reconstruction quality of our proposed transformer-based reconstruction using previous scans (Enhanced [Ours, Transformers]) and an atlas with the non-enhanced reconstructions (Non-enhanced [Ours, E2EVarNet]). The evaluation metrics, including SSIM, NMSE, and PSNR, were computed across various acceleration factors (R5, R10, R15, R20).
The results demonstrate that enhanced reconstructions using previous scans consistently outperforms the non-enhanced and enhanced reconstructions using the atlas as a prior instead of a subject-specific previous scan.

\subsection{Expert Readers Image Quality Assessment}
\begin{figure}[!t]
\centering{\includegraphics[width=0.4\textwidth]{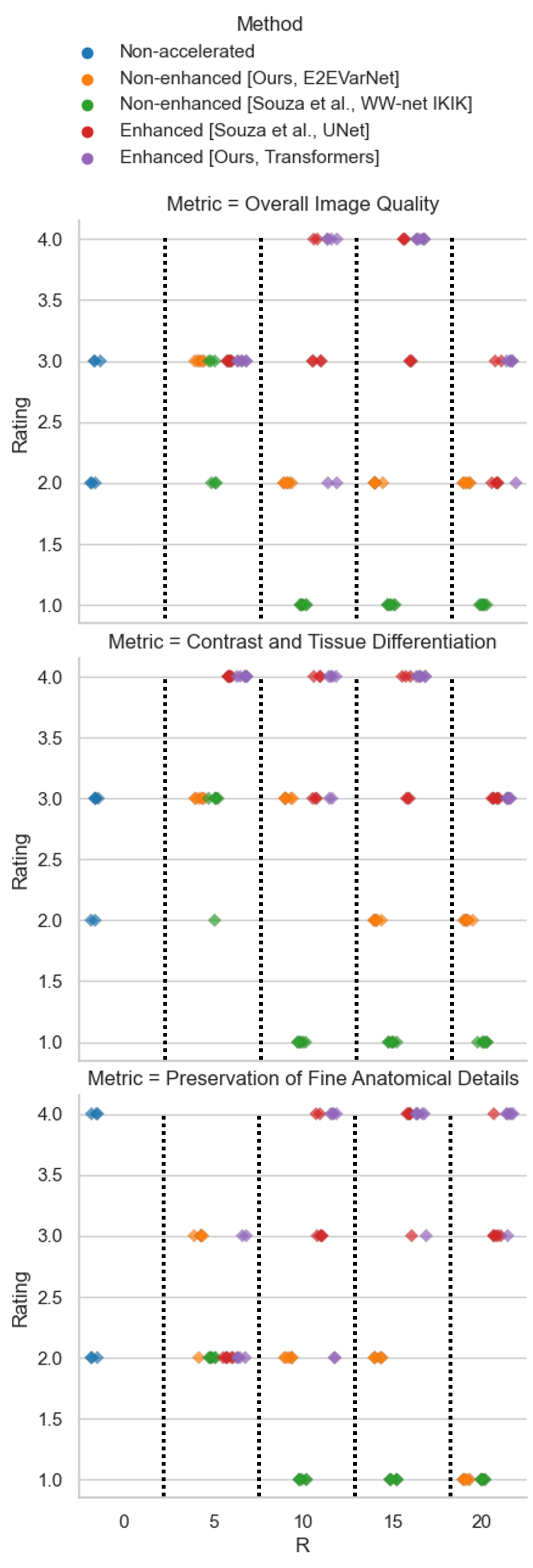}}
\caption{Expert evaluation of image quality across different acceleration factors (R) for various reconstruction methods. Metrics include Overall Image Quality, Contrast and Tissue Differentiation, and Preservation of Fine Anatomical Details, scored on a 4-point scale. The x-axis represents acceleration factors (R), and the y-axis represents the ratings.}
\label{fig:figure8}
\end{figure}

The results of qualitative evaluation performed by expert readers are provided in Figure \ref{fig:figure8}.

\subsubsection{Overall Image Quality}

The non-accelerated images received a mean rating of 2.5 for \textit{Overall Image Quality}. At R5, the Non-enhanced [Ours, E2EVarNet] method received a mean rating of 3.0. The Non-enhanced [Souza et al., WW-net IKIK] method received ratings ranging from 2 to 3, with a mean of 2.5. Both enhanced methods, Enhanced [Ours, Transformers] and Enhanced [Souza et al., UNet], received a consistent mean rating of 3.0.

At R10, the Non-enhanced [Ours, E2EVarNet] method consistently received a rating of 2.0. The Non-enhanced [Souza et al., WW-net IKIK] method received a rating of 1.0 across all evaluations. The Enhanced [Ours, Transformers] method had ratings ranging from 2 to 4, with a mean of approximately 3.33. The Enhanced [Souza et al., UNet] method received ratings of 3 and 4, resulting in a similar mean rating of 3.33.

At R15, the Non-enhanced [Ours, E2EVarNet] method maintained a consistent rating of 2.0. The Non-enhanced [Souza et al., WW-net IKIK] method remained at a rating of 1.0. The Enhanced [Ours, Transformers] method achieved a perfect rating of 4.0 across all evaluations. The Enhanced [Souza et al., UNet] method received ratings of 3 and 4, with a mean of 3.5.

At R20, the Non-enhanced [Ours, E2EVarNet] method continued with a rating of 2.0. The Non-enhanced [Souza et al., WW-net IKIK] method remained at 1.0. The Enhanced [Ours, Transformers] method received ratings of 2 and 3, with a mean of approximately 2.83. The Enhanced [Souza et al., UNet] method received ratings ranging from 2 to 3, resulting in a mean rating of 2.33.

\subsubsection{Contrast and Tissue Differentiation}

The non-accelerated images received mean ratings of approximately 2.67 for \textit{Contrast and Tissue Differentiation}. At R5, the {Non-enhanced [Ours, E2EVarNet]} method had a mean rating of 3.0. The {Non-enhanced [Souza et al., WW-net IKIK]} method had ratings between 2 and 3, with a mean of 2.5. Both enhanced methods achieved a perfect mean rating of 4.0.

At R10, the {Non-enhanced [Ours, E2EVarNet]} method received a consistent rating of 3.0. The {Non-enhanced [Souza et al., WW-net IKIK]} method remained at a rating of 1.0. The {Enhanced [Ours, Transformers]} method had ratings of 3 and 4, with a mean of approximately 3.67. The {Enhanced [Souza et al., UNet]} method received ratings of 3 and 4, resulting in a mean of 3.5.

At R15, the {Non-enhanced [Ours, E2EVarNet]} method had a consistent rating of 2.0. The {Non-enhanced [Souza et al., WW-net IKIK]} method remained at 1.0. The {Enhanced [Ours, Transformers]} method maintained a perfect rating of 4.0. The {Enhanced [Souza et al., UNet]} method had ratings of 3 and 4, with a mean of 3.5.

At R20, the {Non-enhanced [Ours, E2EVarNet]} method received a rating of 2.0. The {Non-enhanced [Souza et al., WW-net IKIK]} method remained at 1.0. Both enhanced methods, {Enhanced [Ours, Transformers]} and {Enhanced [Souza et al., UNet]}, received consistent ratings of 3.0.

\subsubsection{Preservation of Fine Anatomical Details}

The non-accelerated images received a mean rating of approximately 2.83 for \textit{Preservation of Fine Anatomical Details}. At R5, the {Non-enhanced [Ours, E2EVarNet]} method had a mean rating of 2.83. The {Non-enhanced [Souza et al., WW-net IKIK]} method had a mean rating of 2.0. The {Enhanced [Ours, Transformers]} method received ratings of 2 and 3, with a mean of approximately 2.33. The {Enhanced [Souza et al., UNet]} method had a consistent rating of 2.0.

At R10, the {Non-enhanced [Ours, E2EVarNet]} method received a consistent rating of 2.0. The {Non-enhanced [Souza et al., WW-net IKIK]} method remained at 1.0. Both enhanced methods received ratings of 3 and 4, with mean ratings of approximately 3.33.

At R15, the {Non-enhanced [Ours, E2EVarNet]} method maintained a rating of 2.0. The {Non-enhanced [Souza et al., WW-net IKIK]} method stayed at 1.0. The {Enhanced [Ours, Transformers]} method received ratings of 3 and 4, with a mean of approximately 3.83. The {Enhanced [Souza et al., UNet]} method had ratings of 3 and 4, resulting in a mean of 3.83.

At R20, the {Non-enhanced [Ours, E2EVarNet]} and {Non-enhanced [Souza et al., WW-net IKIK]} methods both received consistent ratings of 1.0. The {Enhanced [Ours, Transformers]} method had ratings of 3 and 4, with a mean of approximately 3.83. The {Enhanced [Souza et al., UNet]} method received ratings of 3 and 4, resulting in a mean of 3.17.

\section{Discussion}

The proposed method, which integrates deep registration and a transformer-based enhancement network, demonstrates superior performance compared to non-enhanced reconstructions and the previously proposed linearly registered enhanced reconstructions. The experimental results, conducted on a longitudinal dataset with rigorous evaluation metrics, show that the proposed approach consistently achieves higher SSIM, PSNR, and lower NMSE values across all acceleration factors (Table \ref{tab:image_enhancement_comparison}). These findings underscore the effectiveness of the deep registration technique in accurately aligning prior subject-specific data with the initial reconstruction, as well as the power of the transformer architecture in capturing long-range dependencies and refining the reconstructed images.

The visual comparison of the reconstructed images further reinforces the superiority of the proposed method in preserving fine anatomical details and reducing artifacts, particularly at higher acceleration factors (Figure \ref{fig:figure3}). This improvement in image quality has significant implications for clinical diagnosis and treatment planning, as it enables faster MRI acquisitions without compromising the quality of the images.

Moreover, the segmentation analysis demonstrates the positive impact of the proposed fast MRI reconstruction approach on downstream tasks (Figure \ref{fig:figure4} and Figure \ref{fig:figure5}). The enhanced reconstructions exhibit higher agreement with the reference segmentation masks, especially in the cortical regions, compared to the non-enhanced reconstructions. This finding is crucial, as accurate segmentation of brain structures is essential for various clinical and research applications, such as studying neurodegenerative diseases, monitoring treatment response, and understanding brain development.

The quantitative evaluation of the segmentation results, using Dice similarity coefficients and volumetric agreement (Figure \ref{fig:figure5} and Figure \ref{fig:figure6}), further validates the benefits of the proposed method. The enhanced reconstructions consistently achieve higher Dice scores and lower percentage errors in volume estimates for key brain regions across all acceleration factors. These results highlight the potential of the proposed approach to maintain accurate volumetric measurements despite faster MRI acquisitions, which is critical for longitudinal studies and clinical decision-making.

In terms of computational efficiency, the deep registration approach using EasyReg significantly reduces the registration time compared to the linear registration method used in previous work. This substantial reduction in registration time, combined with no decrease in image quality, makes the proposed approach more feasible for real-time clinical applications, addressing the challenges associated with prolonged MRI acquisition times.

The findings of this study reveal that higher acceleration and enhanced MRI images were rated higher in quality than non-accelerated images by the evaluating expert readers. This challenges the conventional assumption that increasing acceleration factors inherently degrade image quality. The {Enhanced [Ours, Transformers]} method consistently received higher ratings across all evaluation criteria, especially at higher acceleration factors (R5, R10, R15, R20). At R15, it achieved perfect mean ratings for \textit{Overall Image Quality} and \textit{Contrast and Tissue Differentiation}, and a high rating for \textit{Preservation of Fine Anatomical Details}. The {Enhanced [Souza et al., UNet]} method also outperformed the non-accelerated images, though its ratings were slightly lower than those of our enhanced method.

The advanced reconstruction algorithms, particularly the transformer-based architecture in our method, may significantly enhance image quality. These techniques likely reduce noise and artifacts, improve image sharpness, and enhance contrast, leading to better visualization of anatomical structures.
Non-accelerated images require longer acquisition times, increasing the likelihood of patient movement and resultant motion artifacts. Accelerated imaging reduces scan times, potentially minimizing these artifacts and improving image quality. The transformer-based architecture may capture complex spatial relationships more effectively than traditional methods, resulting in higher-quality reconstructions. This could explain why enhanced images were rated higher than non-accelerated images.

\subsection{Limitations}
While the proposed deep-learning-based MRI reconstruction framework demonstrates significant improvements in image quality, computational efficiency, and downstream task performance, there are several limitations to consider:

1. Dataset: The study utilized a relatively small dataset consisting of 79 T1-weighted brain MRI scans from presumably healthy subjects. Although the dataset included longitudinal exams, the limited sample size and focus on healthy individuals may not fully represent the variability encountered in clinical practice. To ensure the generalizability and robustness of our proposed method, it is essential to validate its performance across larger and more diverse datasets, encompassing patients with various neurological conditions and a broader range of age groups. Particularly, assessing the method's efficacy in patients who develop new pathologies after the initial reference scan is crucial. For instance, in multiple sclerosis (MS) patients, the accurate detection of new or enlarging lesions is vital for monitoring disease progression and treatment efficacy. Evaluating whether our accelerated MRI reconstruction technique maintains the visibility of such lesions is imperative to confirm its clinical applicability in this context. Therefore, future research should focus on testing the method in populations where new pathological developments are anticipated, such as MS patients, to ensure that critical diagnostic information is preserved.

2. MRI Contrasts: The current study focused solely on T1-weighted brain images. While T1-weighted scans are commonly used in clinical practice, other MRI contrasts, such as T2-weighted, fluid-attenuated inversion recovery (FLAIR), and diffusion-weighted imaging (DWI), provide complementary information crucial for diagnosing and monitoring various neurological diseases. Evaluating the performance of the proposed method on different MRI contrasts and exploring its potential for multi-contrast reconstruction would further enhance its clinical applicability.

3. Anatomical Regions: The proposed framework was developed and evaluated specifically for brain MRI reconstruction. However, MRI is widely used for imaging other anatomical regions, such as the spine, musculoskeletal system, abdomen, and pelvis. Each of these regions presents unique challenges in terms of image quality, motion artifacts, and anatomical variability. To expand the potential applications of the proposed method, future studies should investigate its performance and adaptability to different anatomical regions.

4. Comparison with Other Methods: While the proposed method outperformed the non-enhanced reconstructions and the previously proposed linearly registered enhanced reconstructions, a comprehensive comparison with other state-of-the-art deep-learning-based MRI reconstruction techniques was not conducted. Future research should benchmark the performance of the proposed framework against a wider range of existing methods to establish its relative effectiveness and identify potential areas for further improvement.

5. The expert readers evaluation involved images from only six subjects, which limits the generalizability of the findings. A larger sample size would provide more robust data. Moreover, only two expert readers evaluated the images. Including more observers would allow for assessment of inter-observer variability and enhance the reliability of the results. The expert readers were blinded to the acceleration status, so despite the experts assigning higher quality scores to some reconstruction methods at higher acceleration factors, this is subjective quality, and these visually pleasing reconstructions may not accurately depict the underlying anatomical structures in the brain.

Addressing these limitations in future studies will provide a more comprehensive understanding of the proposed method's capabilities, limitations, and potential for clinical translation. By expanding the scope of evaluation, incorporating clinical validation, and optimizing the framework for real-world deployment, the proposed deep-learning-based MRI reconstruction approach can be further refined to meet the diverse needs of clinical practice and research, for instance, our proposed framework demonstrates significant advantages in scenarios involving patients prone to motion artifacts, such as those with dementia. These individuals often struggle with remaining still during scans, making fast acquisitions crucial for obtaining diagnostic-quality images.

\section{Conclusion}
In conclusion, the proposed deep-learning-based MRI reconstruction framework, leveraging deep registration and a transformer-based enhancement network, represents an advancement in the field. The improved reconstruction quality, faster registration process, and positive impact on downstream tasks demonstrate the potential of this approach to accelerate MRI examinations while maintaining high image quality. The integration of this method into clinical workflows could lead to increased patient throughput, reduced wait times, and improved patient care. Furthermore, the enhanced image quality and accurate volumetric measurements enabled by this approach may facilitate more precise diagnosis, treatment planning, and research outcomes. Future work should focus on validating the method on larger and more diverse datasets, exploring its generalizability to other anatomical regions and imaging modalities, and assessing its impact on clinical decision-making and patient outcomes.

\section*{Acknowledgment}

Dr. Roberto Souza thanks NSERC for ongoing operating support for this project (RGPIN/2021-02858) and support from the NSERC Alliance–Alberta Innovates Advance Program.

\bibliographystyle{elsarticle-num}
\bibliography{REF}

\end{document}